\documentclass[letterpaper, 10 pt, conference]{ieeeconf}

\IEEEoverridecommandlockouts
\usepackage{amsmath,amsfonts}
\usepackage[algo2e,ruled,vlined,linesnumbered,resetcount,norelsize]{algorithm2e}
\usepackage{algorithm}
\usepackage{array}
\usepackage[caption=false,font=normalsize,labelfont=sf,textfont=sf]{subfig}
\usepackage{textcomp}
\usepackage{stfloats}
\usepackage{url}
\usepackage{comment}
\usepackage{verbatim}
\usepackage{graphicx}
\usepackage{cite}
\usepackage{bm}
\usepackage{mathtools} 
\usepackage{caption}

\usepackage{amssymb}  

\usepackage{amsthm}
\usepackage{xcolor}
\usepackage{accents}
\usepackage{mathrsfs}
\usepackage[normalem]{ulem}
\usepackage{multirow}
\usepackage{tabularx,booktabs}

\newtheorem{remark}{Remark}
\newtheorem{assumption}{Assumption}

\SetCommentSty{mycommfont}

\title{\LARGE \bf
Bi-CL: A Reinforcement Learning Framework for Robots Coordination Through Bi-level Optimization
\author{Zechen Hu, Daigo Shishika, Xuesu Xiao, and Xuan Wang
\thanks{Work supported by  Army Research Office (W911NF-22-2-0242) and NSF (2332210). George Mason University. {\tt\scriptsize \{zhu3, dshishik, xiao, xwang64\}@gmu.edu}. 
}
}
}
\begin{document}

\maketitle

\begin{abstract}
    In multi-robot systems, achieving coordinated missions remains a significant challenge due to the coupled nature of coordination behaviors and the lack of global information for individual robots.
    To mitigate these challenges, this paper introduces a novel approach, Bi-level Coordination Learning (Bi-CL), that leverages a bi-level optimization structure within a centralized training and decentralized execution paradigm. Our bi-level reformulation decomposes the original problem into a reinforcement learning level with reduced action space, and an imitation learning level that gains demonstrations from a global optimizer. Both levels contribute to improved learning efficiency and scalability. We note that robots' incomplete information leads to mismatches between the two levels of learning models. To address this, Bi-CL further integrates an alignment penalty mechanism, aiming to minimize the discrepancy between the two levels without degrading their training efficiency. We introduce a running example to conceptualize the problem formulation and apply Bi-CL to two variations of this example: route-based and graph-based scenarios. Simulation results demonstrate that Bi-CL can learn more efficiently and achieve comparable performance with traditional multi-agent reinforcement learning baselines for multi-robot coordination.



    
\end{abstract}

 \section{Introduction}


Multi-robot systems have extensive applications in various engineering fields, but their deployment relies on scalable algorithms that enable robots to make coordinated and sequential decisions using local observation~\cite{rizk2019cooperative,wang2021distributed}. 
To this end, Centralized Training with Decentralized Execution (CTDE)~\cite{lowe2017multi} emerges as a promising approach, offering a balanced framework for coordinating multiple robots through centralized learning processes that guide decentralized operational decisions. However, the effectiveness of CTDE faces significant challenges as the dimensionality of action spaces expands and each robot only has local observation of the system. These challenges complicate the training process and potentially impedes practical deployment.

In this context, we note that multi-robot cooperative missions often exhibit inherent hierarchical structures, allowing them to be decomposed into high-level and low-level tasks~\cite{yan2013survey}. For instance, rescue missions utilize teams of robots to search large or hazardous areas~\cite{ xiao2021autonomous, liu2021team}, where a high-level planning task may focus on area coverage, and a low-level task may address navigation and obstacle avoidance. A similar scenario arises from warehouse automation~\cite{lian2022spatio}, where high-level tasks involve managing robot fleets for picking, packing, and sorting, and low-level tasks ensure precise and safe navigation in densely populated environments.
Nevertheless, in most cases, the decomposed problems are internally coupled by state dependence, which can not be solved independently. Bi-level optimization~\cite{liu2021investigating} presents a solution to this issue, with the capability of enhancing learning efficiency and stability while maintaining the explicit connections between the two levels of problems. However, despite the abundance of research on addressing such coupling in static optimization problems for both single-agent~\cite{biswas2019literature} or multi-agent~\cite{zhang2020bi} cases, there remains a scarcity of studies tailored for multi-agent reinforcement learning (MARL) under CTDE with robots' local observation. This gap motivates our research to develop new approaches that leverage the benefits of bi-level optimization for reinforcement learning applications in multi-robot systems, thereby enhancing their deployment efficiency across a variety of complex and dynamic environments.

\noindent\textbf{Statement of contribution:}
The contributions of this paper include 
(i) the formulation of a bi-level Decentralized Markov Decision Process (Dec-MDP) for multi-robot coordination; 
(ii) a Bi-level Coordination Learning (Bi-CL) framework with a novel alignment mechanism to integrate multi-agent reinforcement learning and imitation learning under the CTDE scheme; (iii) simulated experiments that verify the effectiveness of the proposed Bi-CL algorithm and a comparison with traditional MARL algorithms to solve multi-robot coordination tasks. 

\begin{figure}
    \centering
    \includegraphics[width=0.46\textwidth]{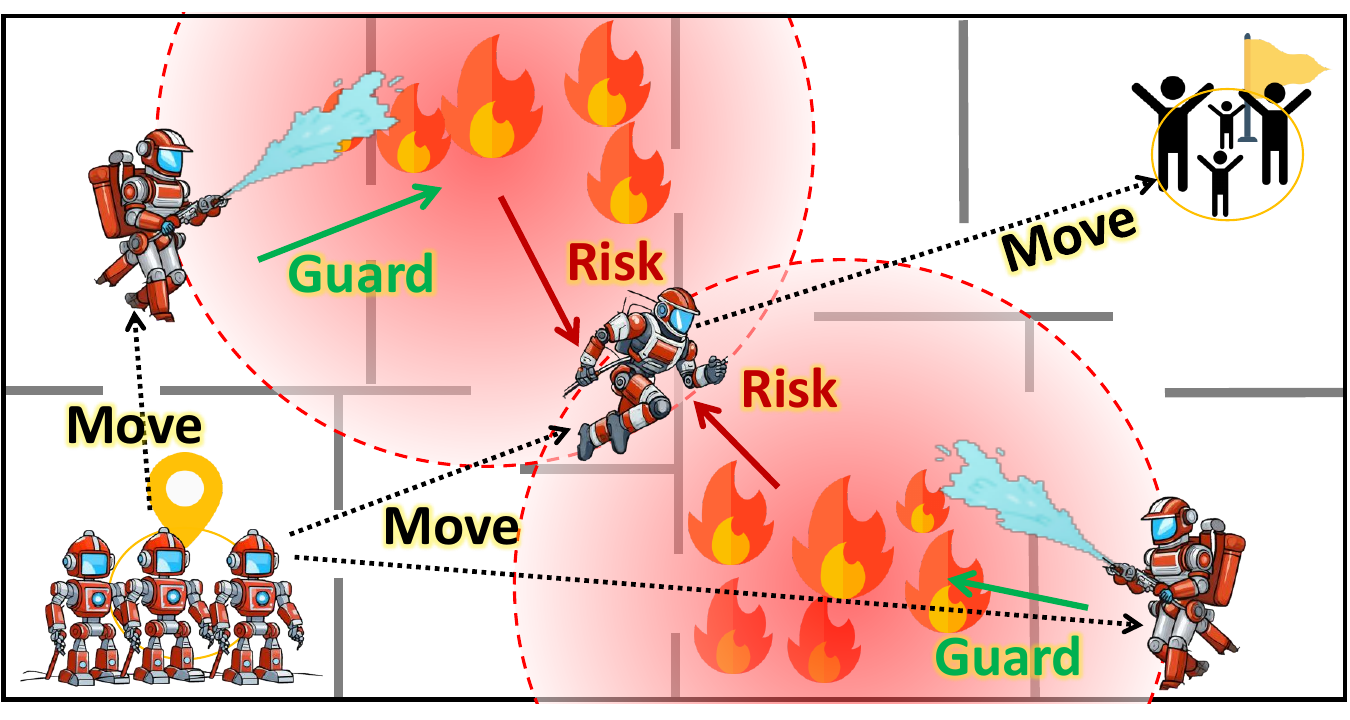}
  \caption{A running example for a firefighting scenario. Robots can simultaneously perform two actions: move (to where) and guard (which adversary). Team reward depends on the risk of fire, which is a coupled function of both actions.}
   \label{fig_example1}
\end{figure}
\noindent\textbf{Running Example:} As visualized in Fig. \ref{fig_example1}, we introduce an example throughout this paper to conceptualize and verify the proposed bi-level approach for learning multi-robot coordination. 
Consider multiple robots traversing an area with multiple adversaries. During their traversal, each robot suffers risk or damage from adversaries accumulated over time, if it enters their impact range. We use robots' states to represent their physical locations. Each robot has two decomposable actions: the action of \textit{move} and the action of \textit{guard}, which are performed simultaneously. 
The \textit{move} action changes the robot's locations in the environment. In each step, the relative positions of robots to adversaries determine a base risk the robot accumulates. The \textit{guard} is an additional action each robot can choose to perform against an adversary, which increases its own risk but reduces the risks that this adversary poses to other robots. 
Our goal is to minimize the accumulated risk and traveling time for all robots to arrive at a target position, given each robot only has local (partial) observation of the system. 

Depending on the weights in performance evaluation, the team may face trade-offs between strategies. For example, each robot may move along the shortest path to reach the target as soon as possible v.s. spending more time to form guard behaviors to reduce team risk. These strategies boil down to robots' interdependent decision-making regarding move (to where) and guard (which adversary) at each time step, making the problem non-trivial. 
On the other hand, as a key idea this paper highlights, the move and guard actions of robots in this scenario exhibit a hierarchical structure. This structure can be leveraged through a bi-level reformulation, which simplifies the problem's complexity while still characterizing the interdependencies between the robots' two types of actions.

The same scenario discussed above applies to a handful of applications for multi-robot coordination, such as firefighting considering fire as adversaries; battlefields considering enemy robots as adversaries; and team car racing or football games considering non-cooperative players as adversaries.

 \section{Literature Review}

\subsection{MARL for Multi-Robot Coordination}
Reinforcement Learning (RL) has gained significant popularity for controlling robotic systems for complex tasks~\cite{kaelbling1996reinforcement}. Using RL for multi-robot systems faces the challenge of increased system dimension and the need for robots to make decisions based on local observations. Addressing these challenges has led to the development of the CTDE~\cite{lowe2017multi} learning scheme, which enables robots to cooperatively learn coordination strategies through centralized training, while the learned policies are local and, thus, can be deployed for decentralized execution.  Examples include MADDPG~\cite{lowe2017multi}, which extends the centralized Deep Deterministic Policy Gradient (DDPG) framework to multi-robot systems. A similar case applies to Proximal Policy Optimization (PPO) algorithm and its generalization MAPPO \cite{yu2022surprising}. While these approaches might be concerned with state dimensionality during centralized training, recent advances, such as Value-Decomposition Networks (VDN), can be leveraged to decompose the joint value function into individual value functions for each robot~\cite{sunehag2017value}. Building on this, QMIX further extends the framework, allowing for a more complex, state-dependent mixing of individual robot's value functions to learn coordinated behaviors~\cite{rashid2020monotonic}. 

These CTDE approaches often rely on end-to-end learning, which cannot leverage the hierarchical structures of robot actions considered in this paper. In addition, their performance and training efficiency may significantly degrade~\cite{gmytrasiewicz2005framework} if each robot's local observation covers only parts of the state space.
Such local observation leads to a Decentralized Markov Decision Process (Dec-MDP), whose complexity is NEXP-complete~\cite{bernstein2002complexity}. In this case, even a slight reduction in the state/action space can have a huge impact on the overall computational complexity.

\subsection{Bi-level Optimization}

Regarding state/action space reduction,
bi-level optimization is a hierarchical mathematical formulation where the solution of one optimization task is restricted by the solution set mapping of another task~\cite{liu2021investigating}.
In recent years, this technique has been incorporated with various machine learning methods for nested decision-making, including multi-task meta-learning~\cite{antoniou2018train}, neural architecture search~\cite{liu2018darts}, adversarial learning~\cite{pfau2016connecting}, and deep reinforcement learning~\cite{hong2020two}. The applications of these bi-level learning approaches span economics, management, optimal control, systems engineering, and resource allocation problems~\cite{wang2022consensus,sinha2017review,mintz2018control,nisha2022bilevel,wang2019scalable}, with a comprehensive review by Liu et al.~\cite{liu2021investigating}. 
Corresponding to the scope of this paper, when solving multi-robot coordination problems, bi-level decomposition has been introduced~ \cite{zhang2020bi,zheng2022stackelberg} for actor-critic MARL to reduce the training burden of each level, thus improving the training efficiency and convergence. 
Le et al.~\cite{le2017coordinated} uses latent variables to introduce a mechanism similar to bi-level optimization that can significantly improve the performance of multi-agent imitation learning.

Nevertheless, a commonality among the mentioned works is the assumption that robots (agents) can access complete state information of the system.
In cases where each robot has only partial information, the incomplete information can lead to a mismatch between the optimal solutions on the two levels. This challenge has motivated our work to incorporate an alignment penalty in order to generalize the bi-level optimization concept under CTDE, taking into account robots' local observations.

 \section{Preliminaries and Problem Formulation}
\subsection{Formulation of a Bi-level Optimization }\label{sec_centralized_bi}
A regular centralized Markov Decision Process (MDP) can be defined by a tuple $(\mathcal{S}, \mathcal{A}, \mathcal{T}, \gamma, {R})$, including state, action, state transition, discount factor, and reward. The goal is to learn a policy $\pi: \mathcal{S}\to\mathcal{A}$ to maximize the expected cumulative reward over the task horizon $T$, i.e.,
\begin{align}
    \max_{\pi} \quad\underset{{\bm{a}}_t \sim \pi(\cdot|{\bm{s}}_t)}{\mathbb{E}}\left[\sum_{t=0}^T \gamma^t R_t\right],  
    \label{eq_c_prob}
\end{align}
where $\bm{a}_t\in\mathcal{A}$ and $\bm{s}_t\in\mathcal{S}$ are the action and state of the system at each step.

Let's assume the actions of the system can be decomposed as $\bm{a}_t=\{{\bm{x}}_t,{\bm{y}}_t\}\in{\mathcal{X}}\times {\mathcal{Y}}$, and the system's state transition depends only on action ${\bm{x}}_t$ such that $\mathcal{T}:\mathcal{S}\times{\mathcal{X}}\to\mathcal{S}$:
\begin{align}\label{eq_trans}
    \bm{s}_{t+1}=\mathcal{T}(\bm{s}_t, {\bm{x}}_t).
\end{align}
However, the reward function depends on both actions:
\begin{align}\label{eq_reward}
    {R}_t=U({\bm{s}}_t,{\bm{x}}_t,{\bm{y}}_t). 
\end{align}
If the structure of problem \eqref{eq_c_prob} satisfies \eqref{eq_trans} and \eqref{eq_reward}, it can be formulated into a bi-level problem:
\begin{subequations}\label{eq_c_prob2}
    \begin{align}\label{eq_bilevel}
    &\max_{\widehat\pi} \quad\underset{{\bm{x}}_t \sim \widehat\pi(\cdot|{\bm{s}}_t)}{\mathbb{E}}\left[\sum_{t=0}^T \gamma^t U({\bm{s}}_t,{\bm{x}}_t,{\bm{y}}^{*}_t({\bm{s}}_t,{\bm{x}}_t))\right], \\
    &~\text{s.t.}\qquad {\bm{y}}^{*}_t({\bm{s}}_t,{\bm{x}}_t)= \arg\max_{{\bm{y}}_t}U({\bm{s}}_t,{\bm{x}}_t,{\bm{y}}_t). \label{eq_opt_bara}
\end{align}
\end{subequations}

In this paper, we make the following assumption: 
\begin{assumption}\label{ass_1}
Problem \eqref{eq_bilevel} is complex such that $\widehat \pi$ is solved using a reinforcement learning (e.g. actor-critic method), whereas problem \eqref{eq_opt_bara} can be solved explicitly and be described as a mapping $f: \mathcal{S}\times {\mathcal{X}}\to{\mathcal{Y}}$.
\end{assumption}

The assumption can be justified by the fact that \eqref{eq_bilevel} is a planning problem involving state transitions and rewards for future $T$ steps, while \eqref{eq_opt_bara} only involves a one-step reward.

\begin{figure}
    \centering
    \includegraphics[width=.45\textwidth]{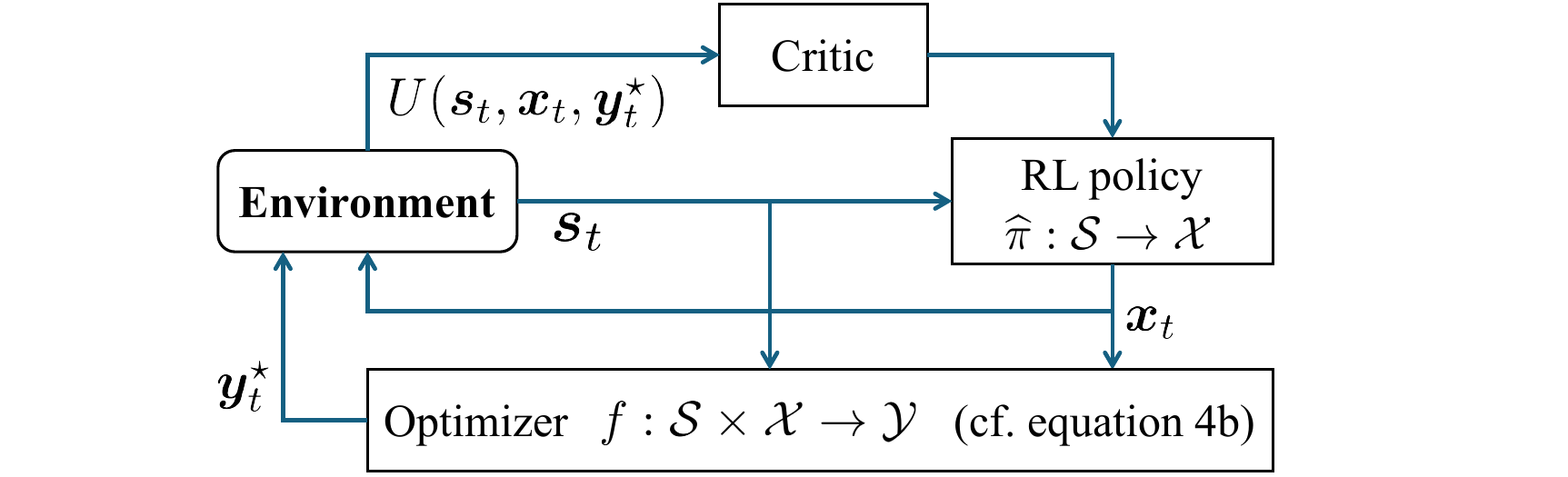}
  \caption{A Centralized Bi-level Optimization for RL.}
   \label{fig_optimizer}
\end{figure}

Under Assumption \ref{ass_1}, for the lower level problem \eqref{eq_opt_bara}, the optimal ${\bm{y}}^{*}_t$ is solvable from $f: \mathcal{S}\times {\mathcal{X}}\to{\mathcal{Y}}$, based on the current $\bm{s}_t$ and the choice of ${\bm{x}}_t$. Bringing this solution back to the upper level \eqref{eq_bilevel}, we only need to solve a reduced-dimension MDP defined by $(\mathcal{S}, {\mathcal{X}}, \mathcal{T}, \gamma, {R})$, to obtain a policy $\widehat \pi: \mathcal{S}\to {\mathcal{X}}$ that generates action ${\bm{x}}_t$. 
During implementation, the policy $\pi$ in problem \eqref{eq_c_prob} can be a composition of 
$\pi=\widehat \pi\circ f$. We learn the policy $\widehat \pi$ to generate ${\bm{x}}_t$ from ${\bm{s}}_t$, then use $f$ to generate ${\bm{y}}_t$ from $\{{\bm{s}}_t,{\bm{x}}_t\}$. Fig. \ref{fig_optimizer} provides a visualization of this bi-level mechanism builds on actor-critic reinforcement leaning.






\subsection{Bi-level Formulation for Multi-robot Coordination Learning with Local Robot Observation}
In this paper, we generalize the bi-level formulation into a multi-robot system with local observation. Consider a multi-robot system with a set of  $\mathbf{n}=\{1,\cdots,n\}$ robots. Suppose the centralized actions and states in \eqref{eq_c_prob} are decomposed into each robot's state and action as $\bm{s}_t = \{s^1_t, \dots, s^n_t\}\in \mathcal{S}$ and $\bm{a}_t = \{a^1_t, \dots, a^n_t\}\in\mathcal{A}$. Assume each robot's action can be further written as $a^i_t = \{{x}^i_t, {y}^i_t\}$, where the action ${x}^i_t$ determines the robot's local state update ${s}^i_{t+1}=\mathcal{T}^i({s}^i_t, {x}^i_t)$.
The multi-robot system shares a global reward determined by all robots' states, and their two actions: ${R}_t=U({\bm{s}}_t,{\bm{x}}_t,{\bm{y}}_t)$.


For many real-world applications, the deployment of a multi-robot system faces communication and sensing constraints, where each agent may only have access to the states of its neighbors through a network, denoted by set $\mathcal{N}_i$. The neighbors' states are a subset of the full system state, denoted by ${s}^{\mathcal{N}_i}_t\subsetneq {\bm{s}}_t$. 
Such local observation turns the problem into a Dec-MDP. Under this constraint, the \textbf{problem of interest} in this paper is to learn a \textit{local} policy $\pi^i$ for each agent such that:
\begin{align}
    \max_{\pi^i} \quad\underset{{a}^i_t \sim \pi^i(\cdot|{s}^{\mathcal{N}_i}_t)}{\mathbb{E}}\left[\sum_{t=0}^T \gamma^t R_t\right]. 
    \label{eq_loca_prob}
\end{align}
While solving Dec-MDP problems is known to be difficult~\cite{bernstein2002complexity}, in this paper, we will generalize the idea of bi-level optimization under the scheme of CTDE and use action space reduction to alleviate the challenges in high-dimensional parameter tuning for MARL.
Note that such a generalization is non-trivial and the main challenge lies in the local observation and local decision-making of each robot using incomplete information. As a consequence, compared with the centralized case, the robots cannot solve an optimization problem like \eqref{eq_opt_bara} to directly determine their secondary action $y^i_t$.




\noindent\textbf{Running Example.} 
We demonstrate how formulation \eqref{eq_loca_prob} aligns with the running example. 
Let state ${s}_t^i$ represent robots' physical locations. The \textit{move} action is associated with ${x}^i_t$, which impacts the robots' state update. The \textit{guard} action is associated with ${y}^i_t$.
The risk function (negative reward) $U$ in each time step depends on the robots' current locations ${s}_t^i$, their move ${x}^i_t$, and how robots are guarding against adversaries, i.e., ${y}^i_t$.
Thus, the scenario satisfies \eqref{eq_trans}-\eqref{eq_reward} and can be reformulated into a multi-robot bi-level optimization problem. All robots only receive states from their neighbors ${s}^{\mathcal{N}_i}_t\subsetneq {\bm{s}}_t$, through an underlying network. 
 \section{Main Approach}

\begin{figure}[t]
    \centering
    \includegraphics[width=0.47\textwidth]{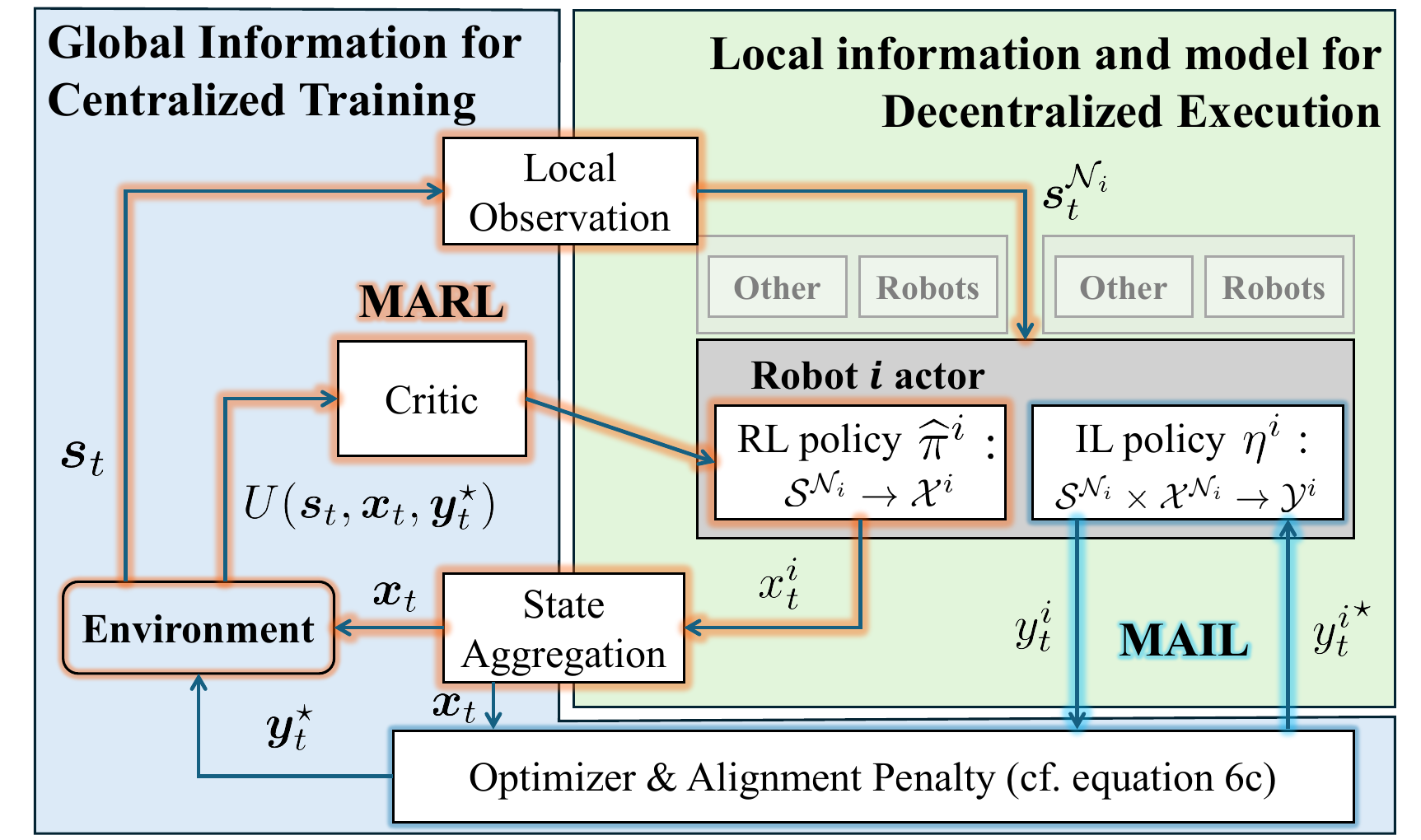}
  \caption{A Bi-level Coordination Learning (Bi-CL) Algorithm: incorporating multi-agent reinforcement learning (MARL) and imitation learning (MAIL), guided by a global optimizer.}
   \label{fig_dist_optimizer}
\end{figure}

As described in Sec. \ref{sec_centralized_bi}, the centralized bi-level formulation leverages the optimizer \eqref{eq_opt_bara} to reduce the dimensionality of the action space, thereby enhancing training efficiency. However, such optimization requires global information, which is not accessible to robots during decentralized execution. Thus, robots cannot locally generate ${y_t^i}^{*}$.
To address this issue, in this section, we introduce a new reformulation to generalize the concept of bi-level optimization to a CTDE  setup. 


To present our approach, we reformulate \eqref{eq_loca_prob} into the following Bi-level Coordination Learning (Bi-CL) problem, $\forall i\in\{1,\cdots,n\}$,
\begin{subequations}\label{eq_d_prob2}
    \begin{align}\label{eq_bilevel_d}
    &\max_{\widehat\pi^i} \quad\underset{{x}^i_t \sim \widehat\pi^i(\cdot|{s}^{\mathcal{N}_i}_t)}{\mathbb{E}}\left[\sum_{t=0}^T \gamma^t U({\bm{s}}_t,{\bm{x}}_t,{\bm{y}}_t^{\star}(\bm{s}_t,{\bm{x}}_t))\right], 
    \\
    &\min_{\eta^i} \quad\underset{{y}^i_t \sim \eta^i(\cdot|{s}^{\mathcal{N}_i}_t,{x}^{\mathcal{N}_i}_t)}{\mathbb{E}}\|{y}^i_t - {{y}^i_t}^*(\bm{s}_t,{\bm{x}}_t)\|^2\label{eq_policy_bara_d}, 
    \\
    &\text{s.t.}~ {\bm{y}}_t^{\star}(\bm{s}_t,{\bm{x}}_t)\!=\! \arg\max_{{\bm{y}}_t}\!\left[U({\bm{s}}_t\!,{\bm{x}}_t\!,{\bm{y}}_t) 
    \underbrace{-c_k \sum_{i=1}^{n}\mathcal{H}_i({y}^i_t, \eta^i))}_{\text{alignment penalty}}\right] \label{eq_opt_bara_d}
\end{align}
\end{subequations}
In equations \eqref{eq_bilevel_d} and \eqref{eq_policy_bara_d}, we aim to train two local policies for each robot: one policy, $\widehat{\pi}^i: \mathcal{S}^{\mathcal{N}_i} \to \mathcal{X}^i$, generates action $x^i_t$; another policy, ${\eta}^i: \mathcal{S}^{\mathcal{N}_i} \times \mathcal{X}^{\mathcal{N}_i} \to \mathcal{Y}^i$, generates action $y^i_t$.
Note that the two policies are coupled since their inputs rely on the outputs of each other. If training two policies simultaneously, such coupling can pose a stability issue and reduce training efficiency. To address this, our idea is to introduce a global optimizer \eqref{eq_opt_bara_d} under the CTDE scheme to decouple and guide the training of both policies, as we explain next and visualized in Fig. \ref{fig_dist_optimizer}.


\noindent\textbf{Bi-level optimization:} First, we introduce optimization problem \eqref{eq_opt_bara_d} to solve ${\bm{y}}_t^{\star}$. If temporarily ignoring the alignment penalty term, the equations \eqref{eq_bilevel_d} and \eqref{eq_opt_bara_d} together represent a CTDE version of the bi-level problem \eqref{eq_c_prob2}.  During the centralized training process, we can utilize global information for all robots' states ${\bm{s}}_t$ and actions ${\bm{x}}_t$ to solve ${\bm{y}}_t^{\star}$. Based on ${\bm{y}}_t^{\star}(\bm{s}_t,{\bm{x}}_t)$, we train the local policy $\widehat\pi^i$ in \eqref{eq_bilevel_d} for each agent's ${{x}}^i_t$ using a \textit{reinforcement learning} scheme. $\widehat\pi^i$ has a \textit{reduced-dimension} action space because ${\bm{y}}_t^{\star}$ is not learned. 

On the other hand, while optimizer \eqref{eq_opt_bara_d} can be used in centralized training to obtain action ${\bm{y}}_t^{\star}$ for all robots, it cannot be used by robots during decentralized execution due to the local observations. Thus, we introduce \eqref{eq_policy_bara_d} to train a local policy $\eta^i$ to generate agent's ${{y}}^i_t$. The policy $\eta^i$ can be trained conveniently using an imitation learning (IL) scheme that uses optimizer \eqref{eq_opt_bara_d} as the \textit{expert demonstration}. 
Here, the imitation loss in \eqref{eq_policy_bara_d} measures the difference between the policy output ${y}^i_t$ using local information and the optimal action solved from optimizer \eqref{eq_opt_bara_d} using global information.

\begin{algorithm2e}[t]
\SetAlgoLined
 \textbf{Initialize} local RL models $\widehat{\pi}^i({s}^{\mathcal{N}_i}|\theta_{\widehat{\pi}^i})$ and IL models $\eta^i({s}^{\mathcal{N}_i}, x^{\mathcal{N}_i}|\theta_{\eta^i})$ for all robots. Initialize centralized critic $Q(\bm{s}, \bm{x}|\theta_Q)$ if needed;  \\
 \textbf{Initialize} alignment penalty co-efficient $c_k$;\\
 \For{$\text{k} = 0$ \KwTo $T_k$}{
    Update coefficient $c_k=\frac{c}{1+e^{-\beta (k-h)}}$;\\
  \For{$t = 0$ \KwTo $T_t$}{
  \For{$i = 1$ \KwTo $n$}{
  Get action: $x^i_t$ by adding random perturbation to ${\widehat{\pi}^i}({s}^{\mathcal{N}_i}_t|\theta_{\widehat{\pi}^i})$;\\
   Get action: $\widehat{y}^i_t = \eta^i({s}^{\mathcal{N}_i}_t,{x}^i_t | \theta_{\eta^i})$;\\
   State update: ${s}^i_{t+1}=\mathcal{T}^i({s}^i_t, {x}^i_t)$;\\
   }
   Solve $\arg\max_{{\bm{y}}_t}\!\left[U(\bm{s}_t, \bm{x}_t, \bm{y}_t) 
    -c_k \sum_{i=1}^n\mathcal{H}_i\right]$ to obtain ${\bm{y}}_t^{\star}$;\\
   Observe reward $R_t = U(\bm{s}_t, \bm{x}_t, \bm{y}_t^{\star});$\\
   Record $(\bm{s}_t, \bm{x}_t, \bm{y}^{\star}_t, R_t, \bm{s}_{t+1})$ to a fixed size buffer $\mathcal{B}$ with first-in-first-out;\\
   Sample $(\bm{s}_{\tau}, \bm{x}_{\tau}, \bm{y}^{\star}_{\tau}, r_{\tau}, \bm{s}_{{\tau}+1})$ from $\mathcal{B}$ as a random minibatch of size $w$;\\
   Update critic network $\theta_{Q}$ using the minibatch, if critic exist;\\
   \For{$i = 1$ \KwTo $n$}{
    Update IL learning model $\theta_{\eta^i}$ using the minibatch;\\
    Update RL learning model: $\theta_{\widehat{\pi}^i}$ using the minibatch;
   }
  }
 }
 \caption{{The Bi-CL Algorithm}}
 \label{alg_1}
\end{algorithm2e}

\noindent\textbf{Alignment of the two policies:} We explain the alignment term in \eqref{eq_opt_bara_d}. Based on the above explanation, the policy $\widehat\pi^i$ is trained assuming the actions ${\bm{y}}_t^{\star}=\{{y^1_t}^*,\cdots,{y^n_t}^*\}$ are optimal for all robots, and the policy $\eta^i$ seeks to generate such optimal actions using only local information. Clearly, during decentralized execution, the optimality of $\eta^i$ can hardly be guaranteed, thus creating a mismatch between the policies $\widehat\pi^i$ and $\eta^i$. To address this, we introduce an alignment penalty term $c_k \sum_{i=1}^{n}\mathcal{H}_i({y}^i_t, \eta^i))$ in \eqref{eq_opt_bara_d}, where $c_k>0$ and
$$\mathcal{H}_i=\underset{\widehat{y}^i_t \sim \eta^i(\cdot|{s}^{\mathcal{N}_i}_t,{x}^{\mathcal{N}_i}_t)}{\mathbb{E}}\|{y}^i_t - {\widehat{y}^i_t}\|^2.$$ 
Here, ${\widehat{y}^i_t}$ is the output of the model $\eta^i$, then $\mathcal{H}_i$ evaluates how well an action ${y}^i_t$ aligns with the policy $\eta^i$. Minimizing this penalty helps to reduce the mismatch between the two local policies. Mathematically, the value of $\mathcal{H}_i$ for each robot equals the loss in \eqref{eq_policy_bara_d}. Hence, $c_k$ is an important coefficient and we shall remark on its choice during the training. If $c_k \to 0$, the reformulation \eqref{eq_d_prob2} disregards the policy mismatch, always opting for the solution that maximizes $U$ to train $\widehat\pi^i$. 
Conversely, when $c_k\to \infty$, the penalty forces the training of $\widehat\pi^i$ using the action chosen by $\eta^i$, making the two policies fully coupled, and the IL \eqref{eq_policy_bara_d} no longer updates. 
To strike a balance, our idea is to make sure the IL model is sufficiently trained before the penalty is applied. For this purpose, we define a modified logistic function:
\begin{align}\label{eq_defck}
    c_k=\frac{c}{1+e^{-\beta (k-h)}}\quad \text{with}\quad c>0, \beta>0, h>0
\end{align}
where $k$ is the training episode. We let $c_k$ start with zero to `jump start' the training of $\widehat\pi^i$ and $\eta^i$ using the optimal ${\bm{y}}_t^{\star}$ for maximizing $U$. Afterwards, $c_k$ is gradually increased to a sufficiently large value $c$ to fine-tune $\widehat\pi^i$, taking into account the output of $\eta^i$ and ensuring alignment between the two policies, $\widehat\pi^i$ and $\eta^i$.
The described training scheme is summarized in Algorithm 1.

\begin{remark}
    In the presentation of Algorithm \eqref{alg_1}, we assume the use of actor-critic-based methods for \eqref{eq_bilevel_d} and a quadratic mismatch function for the imitation loss \eqref{eq_policy_bara_d}. However, the proposed algorithm is also applicable to general CTDE setups when using other reinforcement learning and imitation learning methods. Furthermore, in handling the local observations of robots, there exist more advanced methods \cite[Sec. 2.2.2]{zhang2021multi} in the literature on Dec-MDPs and POMDPs that leverage the memory (historic states) of robots to learn more powerful policies. In this paper, we only use simple memory-less policies, but these more advanced methods are also compatible with Algorithm \eqref{alg_1}.
\end{remark}

 \section{Numerical Results}

In this section, we employ two variants of the running example to verify the proposed algorithm. We use simulated experiments to demonstrate: (i) the effectiveness of the proposed bi-level learning schemes, in particular, the alignment penalty term; (ii) the advantage of the proposed approach in terms of training efficiency compared with alternative MARL algorithms.

Through this section, we use the accumulated reward \eqref{eq_loca_prob} to evaluate the proposed Bi-CL algorithm. To avoid ambiguity, we distinguish the following metrics:

\noindent \textit{\textbf{RL-Reward}}: refers to the reward of the reinforcement learning policy $\widehat{\pi}^i$ when solving \eqref{eq_bilevel_d} during centralized training.

\noindent \textit{\textbf{T-Reward}}: refers to an average reward obtained by executing both the learned reinforcement learning policy $\widehat{\pi}^i$ and imitation learning policy $\eta_i$ in the environment for thirty times. 

\noindent \textit{\textbf{R-Gap}}: refers to the subtraction between the RL-Reward and the T-Reward.

\begin{remark}
    Note that RL-Reward and T-Reward are different because they use different ways to choose robots' action $y_t^i$. RL-Reward uses an `optimal' solution $\bm{y}_t^{\star}=\{{y^1_t}^{\star},\cdots, {y^n_t}^{\star}\}$ of \eqref{eq_opt_bara_d}; T-Reward uses output of the learned policy $\eta_i$. Only the T-Reward reflects the true performance of the policies because it is achievable during decentralized execution, and higher means better. Since R-Gap evaluates the difference, smaller means better.
\end{remark}


\subsection{Coordinated Multi-robot Route Traversal}
\begin{figure}[h]
    \centering
    \includegraphics[width=0.45\textwidth]{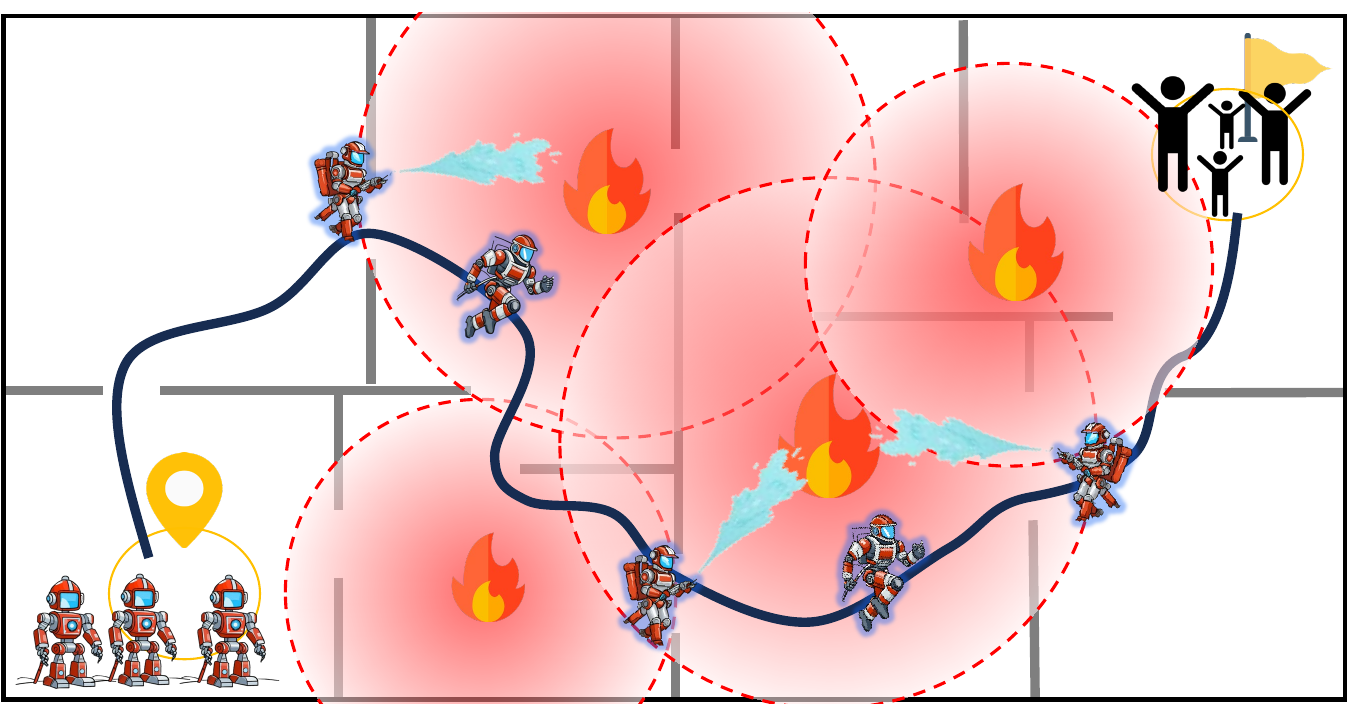}
    \caption{Running Example (a): all robots travel along a route.}
    \label{fig:example_a}
\end{figure}

\begin{figure*}[t]
    \centering
    \includegraphics[width=\textwidth]{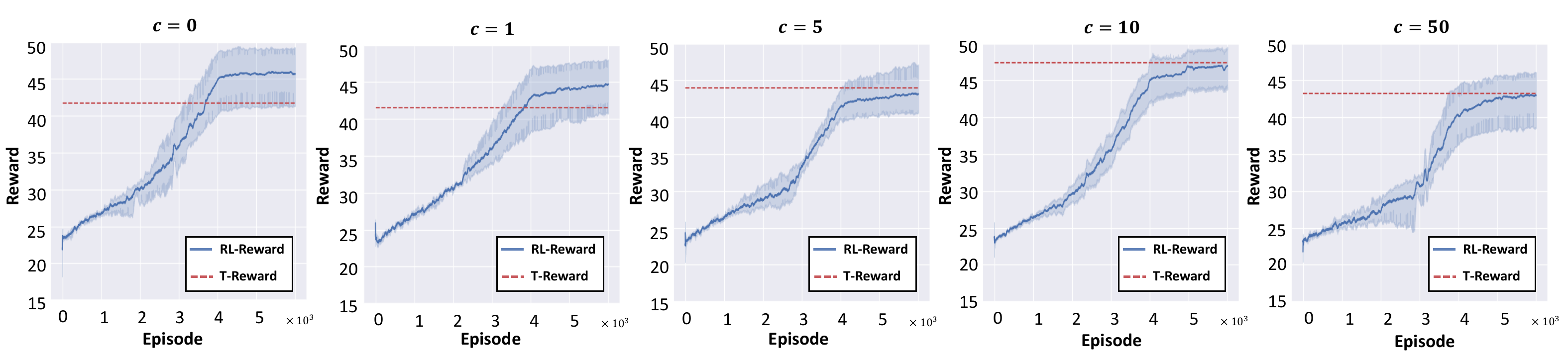}
    \caption{Comparison of cumulative reward for different alignment penalties with four robots and four adversaries. The height of red dash lines determines implementation performance.}
    \label{fig:reward_ct}
\end{figure*}

\noindent\textbf{Running Example (a):}  We introduce a variant of the running example, by assuming all robots travel continuously through a route. Each robot can only observe a subset of full system states through an underlying communication network. Fig. \ref{fig:example_a} provides a visualization of the environment. The testing environment we use is a mathematical abstraction that may not exactly follow its physical layout and the number of adversaries may change.

\noindent~$\bullet$
Each robot $i$ has a continuous \textbf{\textit{move action}} (velocity) $x_t^i \in [-v_{\text{max}}, v_{\text{max}}]$, and a discrete \textbf{\textit{guard action}} $y_t^i \in\mathcal{M}$ where $\mathcal{M}$ is the set of all adversaries. The robot $i$'s \textbf{\textit{state transition}} (position) follows dynamics $s^i_{t+1}=s^i_{t}+x_t^i$. These align with definition \eqref{eq_trans}. 

\noindent~$\bullet$ 
Suppose adversary $j\in\mathcal{M}$ has an impacted area $\mathcal{B}_j$.
Each time-step, if $s^i_{t}\in\mathcal{B}_j$, robot $i$ accumulates a cost $c_j(s^i_{t})$. Besides, for any robot $k$, if $s^k_{t}\in\mathcal{B}_j$, it can perform guard against adversary $j$, i.e., $y^k_t=j$. The guarding effect is characterized by a discount factor on the costs created by adversary $j$:
\begin{align*}
    \alpha^{k,j}_t(x^k_t,y^k_t)=
    \begin{cases}
        1- \beta \frac{(v_{\max}-x^k_t)}{v_{\max}}& y^k_t=j,\\
        1&\text{otherwise}.
    \end{cases}
\end{align*}
Such a discount is more effective when the robot admits at a lower moving velocity $x_t^i$.
The total \textbf{\textit{team cost}} in each step is defined as: 
\begin{align*}
    R_t\!=\! U({\bm{s}}_t,{\bm{x}}_t,{\bm{y}}_t)\!=\!-\!\sum_{i=1}^n\!\sum_{j=1}^m \left[\prod_{k=1}^n \!\alpha^{k,j}_t\!(x^k_t,y^k_t)c^{i,j}_t(s^i_{t})\!\right]\!-\!\delta,
\end{align*}
where $\delta$ is a constant time penalty. A one-time positive reward is added when all robots arrive at the target. The definition of $U$ aligns with definition \eqref{eq_reward}.

\noindent~$\bullet$ The goal is to minimize the team accumulated cost in the form of \eqref{eq_loca_prob} before all robots arrive at the target position. 

The alignment of the problem with \eqref{eq_trans} and \eqref{eq_reward} allows it to be reformulated into \eqref{eq_d_prob2} and then solved by our Bi-CL. 

\medskip
\noindent\textbf{Effectiveness of Alignment Penalty:} 
We first implement the proposed Bi-CL in an environment with 4 robots and 4 adversaries (fire). 
In Bi-CL, the reinforcement learning of $\widehat{\pi}^i$ uses an independent actor in each robot with centralized critic~\cite[Sec. 2.4]{xiao2022asynchronous}, following the structure of MADDPG~\cite{lowe2017multi} but over reduced action space.
Each robot only observes part of the global information, i.e., its own state and the states of a subset of other robots in the system
The cost function $c_j$ is an affine function that depends on the closeness between the robot and the center of the adversary. 

Fig. \ref{fig:reward_ct} illustrates the RL-Reward curves and T-Reward for different $c_t$ by setting $c=0, c=1, c=5, c=10, c=50$, all with $\beta= 2e-3$ and $h=3000$. This, together with \eqref{eq_defck}, makes $c_k$ gradually increase in an `S' shape from $0$ to $c$. 
Upon examining the results, it is evident that the R-Gap is large when $c$ is small, and the gap diminishes as $c$ increases. This verifies the effectiveness of the proposed alignment penalty term in \eqref{eq_opt_bara_d}, as it motivates the policy $\widehat{\pi}_t^i$ to be tuned using a ${y_t^i}^{\star}$ that is closer to the output of $\eta_t^i$. 
It is also important to note that although the MARL training reward appears high when $c$ is small, this is misleading and unattainable in decentralized execution due to the mismatch between $\widehat{\pi}_t^i$ and $\eta_t^i$. The attainable reward (T-Reward) is much lower with smaller $c$.

\begin{table}[b]
    \centering
    \caption{Average reward per episode of different $c_k$ values}
    \small
    \setlength{\tabcolsep}{2.5pt} 

    \begin{tabular}{ccccccc}
        \toprule
        \multirow{2}{*}{Environments} & \multirow{2}{*}{Metric} & \multirow{2}{*}{\(c_k=0\)} & \multicolumn{4}{c}{\(\beta=2e^{-3}\)}\\ 
         \cmidrule(rl){4-7} &  &  & \(c=1\) & \(c=5\)& \(\bm{c=10}\)& \(c=50\)\\
        \midrule 
        \multirow{2}{*}{\textit{N3\_M3}}
        & T-Reward & 51.83 & 53.01 & \textbf{53.44} & 53.17 & 52.13 \\
        & R-Gap & 1.34 & 0.73 & 0.26 & -0.45 & \textbf{-0.23} \\
        \midrule
        \multirow{2}{*}{\textit{N4\_M4}}
        & T-Reward & 42.10 & 42.03 & 44.05 & \textbf{47.38} & {43.56} \\
        & R-Gap & 3.98 & 1.55 & -0.87 & -0.39 & \textbf{-0.11} \\
        \midrule
        \multirow{2}{*}{\textit{N5\_M4}}
        & T-Reward & 56.31 & 58.27 & 59.68 & \textbf{60.08} & 59.46 \\
        & R-Gap & 4.79 & 4.45 & 1.17 & \textbf{-0.04} & -0.15 \\
        \midrule
        \multirow{2}{*}{\textit{N5\_M4}*}
        & T-Reward & 50.67 & 53.39 & 54.09 & \textbf{54.60} & 54.20 \\
        & R-Gap & 6.54 & 3.84 & 1.82 & -0.59 & \textbf{0.27} \\
        \bottomrule
    \end{tabular}
    \label{table_results_cts}
\end{table}

Finally, we observe in the last plot of Fig. \ref{fig:reward_ct} that a very large $c=50$ may negatively impact the training performance. This is due to two reasons. First, as we discussed in the algorithm development, larger $c_k$ reduces the training efficiency of the imitation learning part of the algorithm. This is reflected by the lower values on both RL-Reward and T-Reward. 
Second, since $c_k$ impacts the computation of ${y_t^i}^{\star}$ during the training process, changing it too aggressively will lead to instability in training. This is evidenced by the middle part of the curve, where increased osculation is observed.

The same test is performed under various environment setups and the results are shown in Table \ref{table_results_cts}. Here, \textit{Na\_Mb} means the environment has $a$ robots and $b$ adversaries. 
Especially, in \textit{N5\_M4*}, each robot further reduces the number of other robots's states it can observe. The results in the table align with our above analysis. In all cases where $c_k=c=0$, i.e., without alignment penalty, the performance is the worst. The R-Gap generally reduces when $c$ grows large. Since T-Reward is the core metric for performance, the best result mostly occurs at $c=10$ with only one exception and the difference is small.
Furthermore, to read the table column-wise, environments with the same number of adversaries, i.e., \textit{M4} are comparable. From \textit{N4\_M4} to \textit{N5\_M4}, the T-Rewards generally increase because more coordination behaviors can be generated. When comparing \textit{N5\_M4*} and \textit{N5\_M4}, the T-Rewards generally decrease due to the reduced sensing distances of the robots. This also leads to a larger R-Gap when $c=0$, necessitating the introduced alignment penalty.

\begin{figure}[t]
    \centering
    \includegraphics[width=0.5\textwidth]{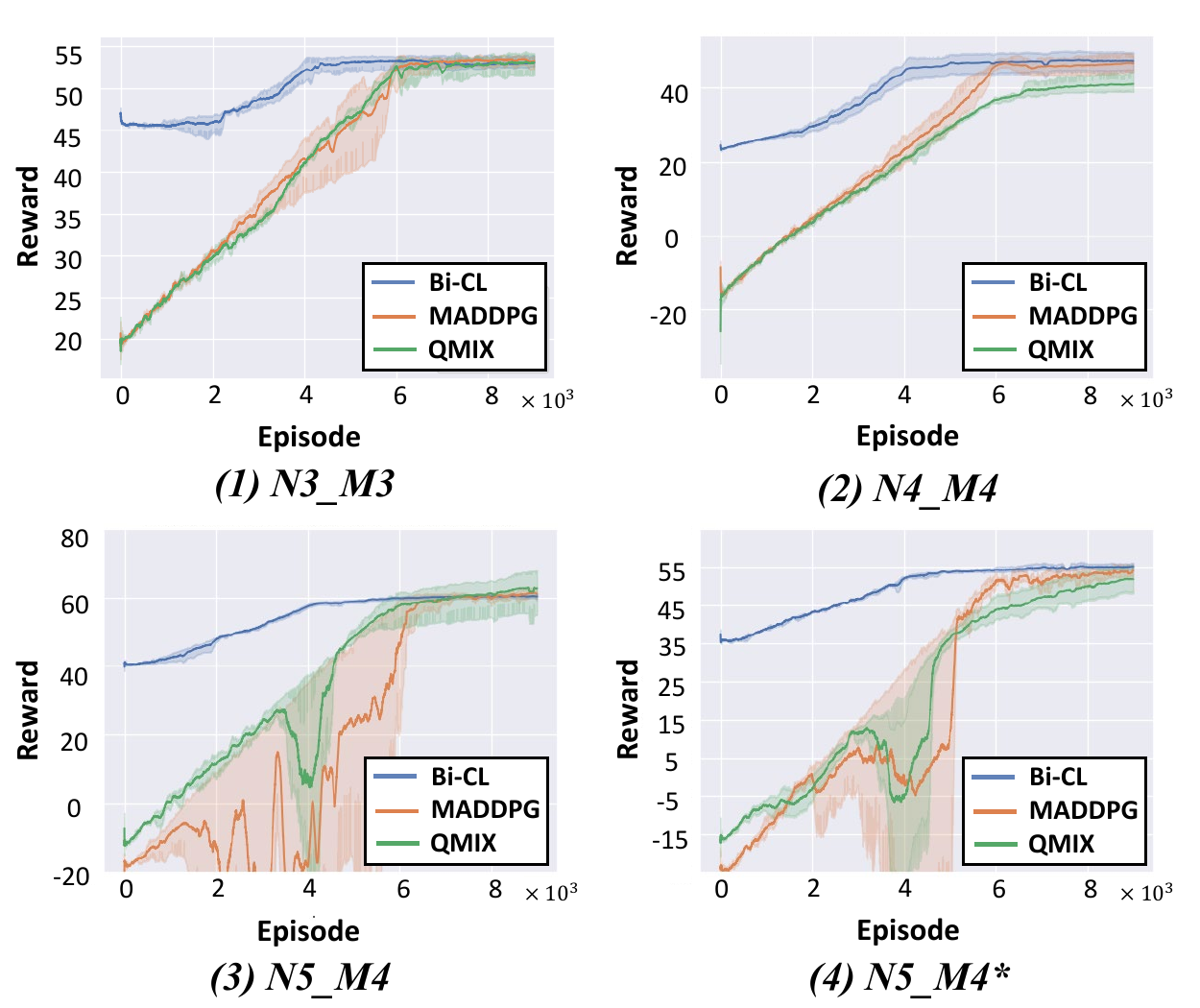}
    \caption{Performance Comparisons in Different Scenarios.}
    \label{fig:algo}
\end{figure}


\medskip
\noindent\textbf{Comparing Training Efficiency with Baselines:} We also compare the proposed Bi-CL algorithm with two well-established MARL algorithms,  MADDPG~\cite{lowe2017multi} and QMIX~\cite{rashid2020monotonic}, respectively. MADDPG and QMIX are implemented in complete action space $\mathcal{A}^i=\mathcal{X}^i\times\mathcal{Y}^i$ of each robot. 
The same learning rates with Bi-CL are used. The comparison is visualized in Fig. \ref{fig:algo} for the four cases in the above discussion. Here, we choose $c=10$ as the R-Gaps are small. Thus, the RL-Reward curve can represent the convergence of our algorithm and can approximate the true reward our algorithm can achieve. It can be observed that in all cases, Bi-CL can achieve similar final rewards compared with baselines, which justifies the effectiveness of the proposed algorithm. The efficiency of Bi-CL is evidenced by its faster convergence speed compared with baselines. 
Furthermore, the starting reward of our algorithm is significantly better because it uses the optimization \eqref{eq_opt_bara_d} to boost and guide the policy training, while other methods are purely based on exploration.
For a similar reason, we observe that both MADDPG and QMIX suffer from training stability issues for complex scenarios, while the proposed Bi-CL does not.

\subsection{Coordinated Multi-robot Graph Traversal}
\begin{figure}[h]
    \centering
    \vspace{-.5em}
    \includegraphics[width=0.45\textwidth]{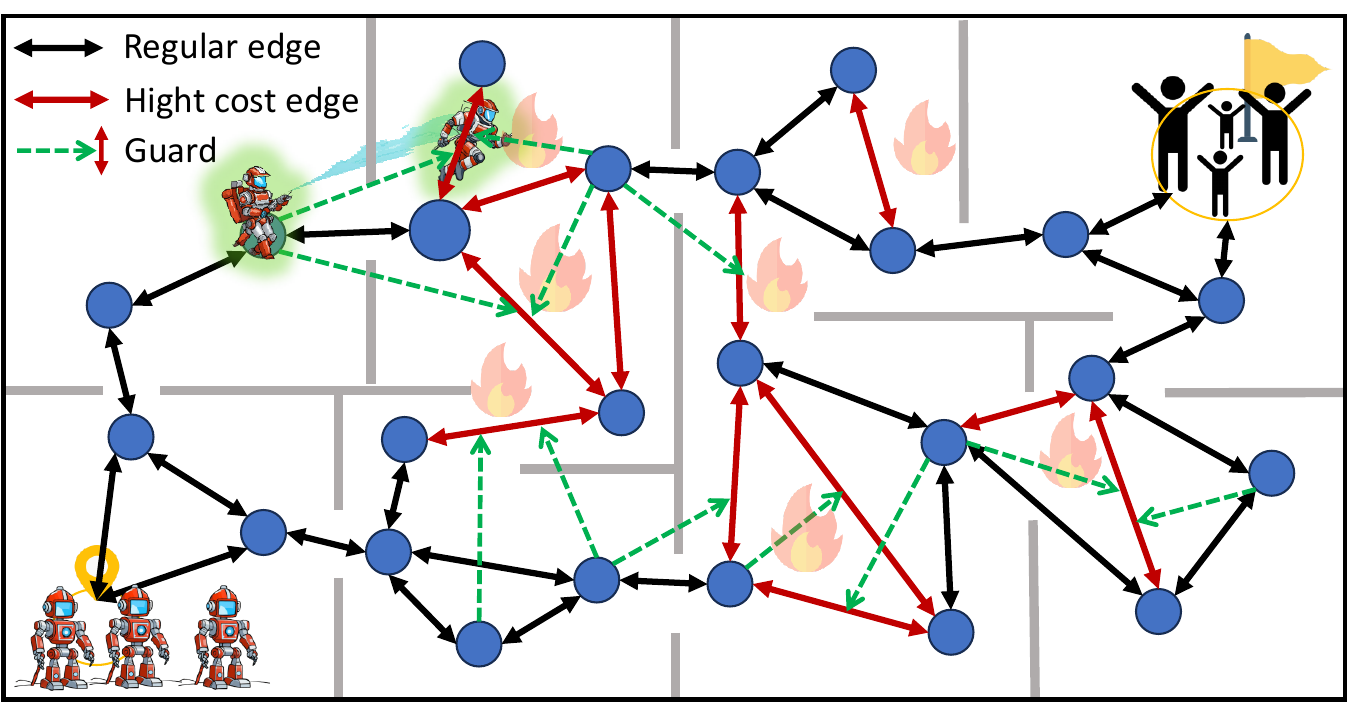}
    \caption{Running Example (b): all robots travel through a graph.}
    \label{fig:example_b}
\end{figure}
\noindent\textbf{Running Example (b):} In the second scenario, we assume all robots travel through a graph environment, which highlights the coordination of robots from a spatial aspect. Fig. \ref{fig:example_b} provides a visualization of the environment; however, the testing environment we use includes randomly generated graphs that may not exactly follow the physical layout. Each robot can only observe a subset of full system states. The environment is characterized by a environment graph describing robot traversability from one node to another. It also features a coordination mechanism (green arrows) describing some robots, when located on certain nodes, can \textit{guard} other robots' edge traversal to reduce their cost, similar to the work by Limbu et al.~\cite{limbu2023team, limbuteam}.

\noindent~$\bullet$
Each robot $i$ has a discrete \textbf{\textit{move action}} $x_t^i\in\mathcal{N}(s^i_{t})$ that moves the robot to a new node in the graph based on its current position $s^i_{t}$. Here, $\mathcal{N}(s^i_{t})$ is the set of adjacent neighboring nodes of robot position  $s^i_{t}$ in the environment graph. The robot $i$'s \textbf{\textit{state transition}} (position) follows $s^i_{t+1}=x_t^i$ subject to $x_t^i\in\mathcal{A}(s^i_{t})$. 
The robot also has a discrete \textbf{\textit{guard action}} $y_t^i\in\mathcal{G}(x_t^i)$, where $\mathcal{G}(x_t^i)$ is the set of edges that can be guarded by a robot at its new position $s^i_{t+1}=x_t^i$ (visualized by green arrows in Fig. \ref{fig:example_b}).
The robots' actions and state transitions align with definition \eqref{eq_trans}.

\noindent~$\bullet$
When robot $i$ travels an edge $\{s^i_{t},x_t^i\}$, i.e., from $s^i_{t}$ to $s^i_{t+1}=x_t^i$, there is an associated edge cost $c(\{s^i_{t},x_t^i\})$. Besides, based on robot coordination, if another robot $k$ can and chooses to guard this edge, i.e., $y^k_t=\{s^i_{t},x_t^i\}\in\mathcal{G}(x_t^k)$, then this edge cost will be discounted by a factor 
\begin{align*}
    \alpha_t^{i,k}(s^i_{t},x_t^i,y^k_t)=
    \begin{cases}
    \alpha^* & y^k_t=\{s^i_{t},x_t^i\}\in\mathcal{G}(x_t^k),
    \\1 & \text{otherwise}.
    \end{cases}
\end{align*}
Based on this, the total \textbf{\textit{team cost}} in each step is defined as: 
\begin{align*}
    R_t\!= \!U({\bm{s}}_t,\!{\bm{x}}_t,\!{\bm{y}}_t)\!=\!-\!\!\sum_{i=1}^n \!\left[\prod_{k=1}^n \!\alpha_t^{i,k}(s^i_{t},x_t^i,y^k_t)c(\{s^i_{t},x_t^i\})\!\right] \!\!-\! \delta
\end{align*}
where $\delta$ is a constant time penalty. A one-time positive reward is added when all robots arrive at the target. The definition of $U$ aligns with definition \eqref{eq_reward}.

\noindent~$\bullet$ The goal is to minimize the team accumulated cost in the form of \eqref{eq_loca_prob} before all robots arrive at the target position. 

The alignment of the problem with \eqref{eq_trans} and \eqref{eq_reward} allows it to be reformulated into \eqref{eq_d_prob2} and then solved by our Bi-CL. 

\begin{table}[ht]
    \centering
    \caption{Comparison of Average Reward and Convergence time for Running Example (b)}
    \small
    \setlength{\tabcolsep}{2.5pt} 

    \begin{tabular}{cccccccc}
        \toprule
        \multirow{ 2}{*}{Graph} & \multirow{ 2}{*}{\begin{tabular}{@{}c@{}}Robots/\\Nodes\end{tabular}}  & \multicolumn{3}{c}{T-Reward} & \multicolumn{3}{c}{Converge ($\times 10^3$)}\\
        \cmidrule(rl){3-5} \cmidrule(rl){6-8} & & Bi-CL & MAPPO & QMIX &  Bi-CL & MAPPO & QMIX \\
        \midrule 
        \multirow{4}{*}{Sparse} 
        & \textit{3/5}
        & \textbf{83.16} & 82.08 & 78.82 & \textbf{3.58} & 4.89 & 5.85 \\
        & \textit{3/10}
        & 64.86 & \textbf{65.39} & 64.97 & \textbf{3.69} & 4.87 & 5.95 \\
        & \textit{5/10}
        & 54.91 & 50.33 & \textbf{55.21} & \textbf{3.75} & 4.96 & 6.14 \\
        & \textit{5/15}
        & \textbf{51.37} & 33.14 & 49.67 & \textbf{3.71} & 5.24 & 6.03 \\
        \midrule
        \multirow{4}{*}{Dense}
        &\textit{3/5}
        & 93.21 & \textbf{96.90} & 94.23 & \textbf{3.63} & 4.95 & 6.19 \\
        & \textit{3/10}
        & \textbf{73.41} & 60.89 & 71.08 & \textbf{3.79} & 5.09 & 6.38 \\
        & \textit{5/10}
        & 66.32 & 54.14 & \textbf{68.08} & \textbf{3.80} & 4.94 & 6.29 \\
        & \textit{5/15}
        & \textbf{60.08} & 47.82 & 58.16 & \textbf{3.85} & 4.94 & 6.44 \\
        \bottomrule
    \end{tabular}
    \label{table_results_alg}
\end{table}

\medskip
\noindent\textbf{Comparing Training efficiency with Baselines:} In this scenario, we compare the proposed Bi-CL algorithm with two MARL algorithms MAPPO~\cite{lowe2017multi} and QMIX~\cite{rashid2020monotonic}, respectively. 
Here, we use MAPPO instead of MADDPG because 
centralized PPO's suitability for this task has been validated by the study in Limbu et al.~\cite{limbuteam}.  
In Bi-CL, the reinforcement learning part is implemented by  QMIX in action space $\mathcal{X}_i$ for each robot. While the MAPPO and original QMIX are implemented in complete action space $\mathcal{A}^i=\mathcal{X}^i\times\mathcal{Y}^i$ of each robot. 
The simulations are configured with the following setups. We experiment with 5, 10, 15 nodes in the environment graph with 3 or 5 robots in two types of graph connectivity, i.e., sparse and dense, for edges and the number of edges that can be guarded. We randomly create 15 environment graphs as our test set. For all approaches, training is terminated when the cumulative reward no longer changes more than $1\%$ for 10 steps. 

The comparison results are presented in Table \ref{table_results_alg}, which verifies the effectiveness of the Bi-CL algorithm for learning coordinated actions among robots within graph-based coordination tasks. Despite variations in the number of robots, graph nodes, or densities of the graph edges, the proposed Bi-CL algorithm delivers comparable results to the baseline algorithms while requiring fewer training episodes to converge. By observing the column of MAPPO, the results tend to fall below those of other methods under conditions of increased environmental complexity. This suggests that policy gradient approaches may only learn a sub-optimal policy solution in these cases.
The QMIX method is more robust in terms of optimality, however, it generally requires more training episodes to converge. In comparison, the proposed Bi-CL method enjoys much more efficient convergence properties. This improvement is mainly due to (i) the reduction of the action space for reinforcement learning, and (ii) the use of optimization \eqref{eq_opt_bara_d} to jump-start the training.

 \section{Conclusion}
We presented a bi-level formulation for multi-robot coordination learning with local observation, wherein robots' state transitions and their cooperative behaviors are abstracted and optimized on different levels, significantly enhancing learning efficiency. 
A key enabler of our Bi-CL algorithm was 
an alignment penalty that enables upper-level learning to account for potential discrepancies arising from local observations in lower-level optimization. We validated our algorithm through a running example, designing two distinct environments: route-based and graph-based. Experimental results demonstrated that our algorithm can effectively learn in both environments, underscoring its versatility and applicability across a diverse set of coordination tasks. We evaluated the performance enhancement of the Bi-CL using different alignment penalty parameters. Comparative analysis with baselines verified the efficiency of our algorithm. 

For future work, we aim to explore the scalability of our Bi-CL to accommodate larger multi-robot systems and more complex environments, further refining the alignment penalty mechanism to enhance its adaptability and efficiency. Moreover, 
we intend to extend our way of handling robots' information loss to effectively manage dynamic, stochastic, and noisy scenarios, thereby enhancing its resilience and performance in unpredictably evolving multi-robot coordination environments. 




\bibliographystyle{ieeetr}
\bibliography{bib}

\end{document}